\documentclass[letterpaper, 10 pt, conference]{ieeeconf}  

\IEEEoverridecommandlockouts                              
\usepackage{censor}
\usepackage{graphicx}
\usepackage{amsfonts}
\usepackage{amsmath}
\usepackage{bm}
\usepackage[hyphens]{url}
\usepackage{xurl}

\usepackage[ruled,vlined]{algorithm2e}
\usepackage{cite}
\usepackage{booktabs}
\usepackage{multirow}
\usepackage{float}

\title{\LARGE \bf
Ensuring Interaction Safety in Multitask Exoskeleton Control: A Simulation-Trained Variable Impedance Framework
}

\author{Muyuan Ma, Houcheng Li, Haotian Zhai, Lijun Han, Xinpan Meng, Xiuze Xia, Long Cheng}

\begin{document}

\maketitle
\thispagestyle{empty}
\pagestyle{empty}

\begin{abstract}
Wearable exoskeletons can augment human physical capabilities during complex activities. However, ensuring adaptation across diverse tasks while guaranteeing interaction safety remains a critical challenge. To address this, a simulation-trained variable impedance control approach with stability guarantees is proposed. First, a simulation-based human-exoskeleton motion data generation pipeline is established, utilizing Proximal Policy Optimization (PPO) to synthesize human muscle activations while the exoskeleton provides direct compensation for human biological joint torques. Subsequently, the generated dataset is used to train a dual modality policy that fuses semantic instructions with proprioceptive history, enabling the prediction of reference trajectories and variable impedance gains for nine different motion tasks. To guarantee safety, the network outputs are constrained by a stability criterion derived from Lyapunov stability theory, which bounds stiffness variations to ensure the asymptotic stability of the coupled system. Experimental results indicate that the proposed framework reduces metabolic cost in real-world scenarios compared with standard baseline methods. These findings suggest the feasibility of the proposed framework for safe, multitask exoskeleton control.
\end{abstract}

\section{Introduction}   
Wearable robots can augment human physical capabilities, offering significant potential to mitigate the risk of musculoskeletal disorders in industrial and rehabilitation settings \cite{1,2}. However, despite success in single-task assistance, widespread deployment in unstructured real-world scenarios remains a challenge \cite{3}. The complexity of daily life requires an exoskeleton to seamlessly adapt to a wide variety of diverse motion tasks, ranging from performing distinct gestures to lifting heavy payloads. Simultaneously, the system must strictly maintain safe physical human-robot interaction.

\begin{figure*}[!t]
	\centering
	\includegraphics[width=\textwidth]{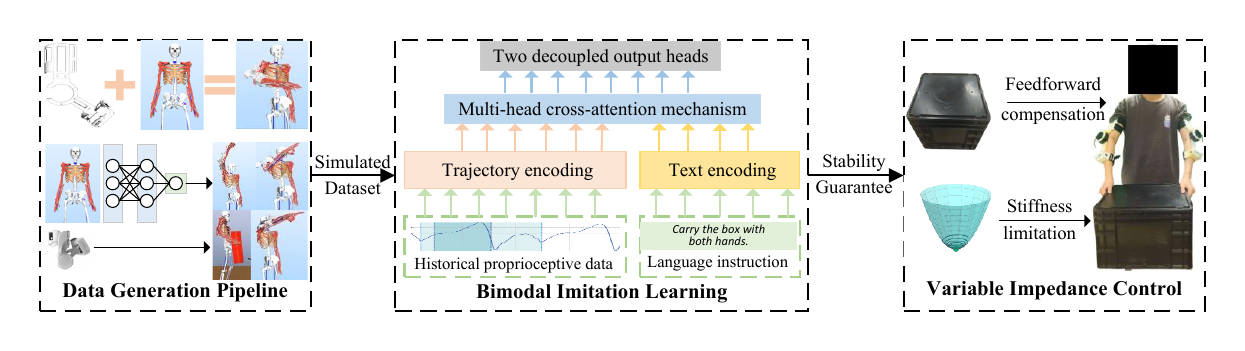}
	\caption{Overview of the proposed simulation-trained variable impedance control framework. This framework consists of three main stages: (1) simulated human-exoskeleton data generation, (2) bimodal imitation learning, and (3) real-world variable impedance control.}
	\label{fig_1}
\end{figure*}

In the literature, task-specific assistance methods for exoskeletons have been extensively investigated. \cite{1.1} developed a CNN-LSTM model that accurately maps electromyography (EMG) signals to future joint trajectories during a multi-joint reaching task. For a specific hand grasping task, \cite{1.3} employed personalized musculoskeletal models to estimate finger joint torques via experimental calibration, enabling dynamic adjustment of assistance levels. \cite{1.10} introduced Flexos, a portable, bilateral active shoulder exoskeleton designed to reduce muscular effort during industrial weight-lifting and carrying tasks. Relying on expert therapists' demonstrations, \cite{1.7} proposed a learning by demonstration framework using Hidden Markov Models (HMMs) to generate trajectories for individual task-specific movements. In addition, \cite{2.6} introduced a deep learning framework that predicts user intention from wearable muscle sensors to provide pneumatic assistance for upper-limb flexion and extension. Nevertheless, these strategies are typically designed for specific tasks, requiring extensive calibration and lacking scalability across diverse movements and interaction scenarios.

To address the lack of scalability in single-task exoskeleton assistance, recent research has increasingly focused on developing multitask assistance frameworks. \cite{1.8} proposed an exoskeletal weight compensation strategy optimized for four types of daily movements. To further enhance adaptability, \cite{3.2} proposed a task-agnostic control strategy based on deep neural networks to provide continuous, multi-joint assistance across 28 diverse and unstructured activities. \cite{2.2} introduced a parametric early motion recognition framework for upper limb mirror rehabilitation to accurately classify multiple therapeutic tasks from partial trajectories. Additionally, \cite{2.9} introduced the ANYexo 2.0 upper-limb exoskeleton featuring a unique bio-inspired kinematic structure to support 15 diverse activities of daily living across all rehabilitation stages. Furthermore, \cite{2.10} proposed a human in the loop adaptation framework that employs a neural task translator to generalize assistance trajectories across multiple complex scenarios. However, these methods rely on real human-exoskeleton interaction datasets, whose collection is time-consuming and requires extensive human trials.

In order to address the challenges of costly real-world data collection, simulation-based data generation and policy training has become an effective paradigm for exoskeleton assistance. For instance, \cite{myo} introduced an automated pipeline to generate computationally efficient musculoskeletal models, leveraging the MuJoCo physics engine to support contact-rich simulations for exoskeleton research. \cite{3.3.3} developed and simulated a hand exoskeleton utilizing an extensive sensor network and PWM control for precise assistance. To train simulated assistive policies, \cite{robustwalking} formulated a deep reinforcement learning framework that employs a decoupled offline strategy to guarantee robust walking control for musculoskeletal-coupled exoskeletons. Furthermore, \cite{3.1} developed an experiment-free sim to real pipeline using reinforcement learning to autonomously synthesize policies for multitask ankle assistance, without requiring any human-in-the-loop trials. Moreover, \cite{3.3} introduced a deep domain adaptation framework that eliminates the need for device-specific labeled data, achieving performance comparable to fully trained models. Despite enabling direct deployment, assistive controllers trained in simulation often lack explicit stability guarantees during real-world human-exoskeleton interaction, resulting in potentially unsafe behaviors.

To leverage the scalability of simulation-based learning for multitask assistance while guaranteeing interaction safety during real-world deployment, a simulation-trained variable impedance control approach with stability guarantees is proposed. The proposed framework is illustrated in Fig. \ref{fig_1}.  In this framework, a simulation environment is utilized to synthesize human-robot interaction data. Leveraging this dataset, a bimodal multitask imitation learning policy is trained to predict reference trajectories and variable impedance gains. To enforce safety, a Lyapunov-based stability criterion is derived to bound stiffness variations, thereby ensuring the stability of the coupled system. The contributions of this article are as follows.

\begin{enumerate}
	\item A data generation pipeline is established for human-exoskeleton coordinated motion in simulation. A detailed musculoskeletal model is integrated with a robotic exoskeleton model to capture realistic biodynamics. PPO is utilized to minimize kinematic tracking errors and biological joint torques, thereby enabling the collection of high-quality interaction data aligned with biomechanical efficiency.
	
	\item A bimodal imitation learning approach is proposed for multitask generalization. The network fuses semantic text embeddings with historical proprioceptive states to encode task intent. This architecture distills the expert policy from simulation, enabling the robot to predict future reference trajectories and variable impedance gains in real-time, adapting to varying dynamic requirements implied by the text input.
	
	\item A stability-guaranteed variable impedance control strategy is formulated using Lyapunov stability theory. A constraint is constructed to bound the variation of the stiffness, ensuring the asymptotic stability of the system. Furthermore, initial physical deployments suggest that the framework can reduce metabolic cost compared to standard baseline methods.
\end{enumerate}

The rest of this article is organized as follows. Section II introduces the simulated human-exoskeleton data generation pipeline. Section III presents the bimodal multitask imitation learning method. Section IV details the design of the variable impedance controller with Lyapunov-based stability guarantees. Section V validates the framework through simulation data analysis, real-world multitask stability evaluations, and metabolic cost assessments. Finally, Section VI concludes this article.

\section{Simulation-Based Human-Exoskeleton Coordinated Motion Generation}
\begin{figure}[!t]
	\centering
	\includegraphics[width=\columnwidth]{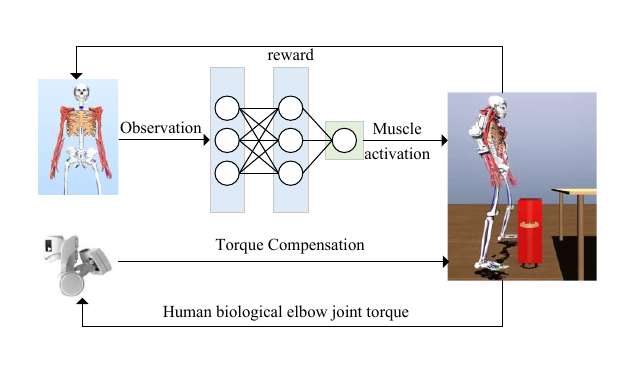}
	\caption{The architecture of the proposed dual-loop co-simulation framework, illustrated using a lifting and carrying task.}
	\label{fig_2}
\end{figure}

To capture the complex biodynamics of human-robot interaction, a high-fidelity simulation environment is constructed within the MuJoCo physics engine. Specifically, a comprehensive dual-arm musculoskeletal model is established, featuring 76 degrees of freedom (DoFs) driven by 126 individual muscle-tendon units. Moreover, a simulation model based on the real exoskeleton is created and fixed to the arms. The integration of these components results in a virtual human-robot coupled system. Within this environment, PPO is employed to drive the musculoskeletal model, enabling the simulation of realistic human motion and different exoskeleton assistance modes. In this process, PPO optimizes muscle activations to accurately track human reference motions, while the exoskeleton provides assistance via proportional control to directly compensate for the biological elbow joint torque. The overall architecture of this co-simulation framework is illustrated in Fig. \ref{fig_2}. The observation space of the PPO policy is defined as
\begin{equation}
s = [q^h_d, q^h, e_h, \dot{q}^h, \tau_{\text{ext}}],
\end{equation}
where $q^h_d$ represents the reference human joint positions derived from public datasets \cite{ref25,ref25-a}; $q^h$ and $\dot{q}^h$ denote the current joint positions and velocities of the musculoskeletal model, respectively; $e_h = q^h_d - q^h$ represents the tracking error between the reference and current joint positions; and $\tau_{\text{ext}}$ is the interaction torque between the human and the exoskeleton. The action space consists of the human muscle activation signals, denoted by $a^h$. The reward function is formulated as a weighted trade-off between the trajectory tracking error and the muscle activation effort, defined as follows:
\begin{equation}
r = -(\left\| q^h_d - q^h \right\|^2 + w \left\| a^h \right\|^2),
\end{equation}
where $w$ is a weighting coefficient that balances the tracking accuracy and the muscle activation penalty. Correspondingly, the assistive torque provided by the exoskeleton, denoted by $\tau_{\text{exo}}$, is defined as a proportional compensation of the human biological joint torque:
\begin{equation}
	\tau_{\text{exo}} = \alpha \tau_{\text{bio}},
\end{equation}
where $\tau_{\text{bio}}$ represents the active biological torque generated by the human musculoskeletal system at the elbow joint, and $\alpha \in [0, 1]$ is the proportional assistance gain. By modulating the value of $\alpha$, different levels of exoskeleton assistance modes can be seamlessly simulated to enrich the interaction dataset.

Upon completion of the training phase, simulation rollouts are conducted to collect interaction data across various assistance ratios $\alpha$. During these rollouts, alongside the semantic text descriptions defining each specific task (e.g., "lift a heavy box"), the interaction torque $\tau_{\text{ext}}$, the reference elbow joint position $q^{\text{elbow}}_d$, and the actual exoskeleton joint position $q^r$ are comprehensively recorded. The human-exoskeleton interaction dynamics are modeled using a linear elastic relationship, formulated as:
\begin{equation}
\tau_{\text{ext}} = K_{\text{interactive}}(q^{\text{elbow}}_d - q^r),	
\end{equation}
where $K_{\text{interactive}}$ is the human-robot interaction stiffness obtained through data fitting.

To guarantee robust tracking performance across diverse operational conditions, nine different motions are selected for training, encompassing various gestures and lifting tasks sourced from public human motion datasets \cite{ref25,ref25-a}. Each motion is evaluated under three discrete assistance ratios ($\alpha \in \{0, 0.5, 1\}$) and three different musculoskeletal model masses (60 kg, 70 kg, and 80 kg). For every specific combination of motion, assistance level, and body mass, a dedicated neural network is optimized using PPO. Consequently, a total of 81 independent policies are trained, ensuring minimal kinematic tracking errors across all simulated scenarios. To support reproducibility and future research, the final simulation dataset has been made publicly available and is accessible via an anonymous link: \url{https://anonymous.4open.science/r/human_exo_mocap_dataset-743B}.

\section{Bimodal Multitask Policy Learning}

\begin{figure}[!t]
	\centering
	\includegraphics[width=\columnwidth]{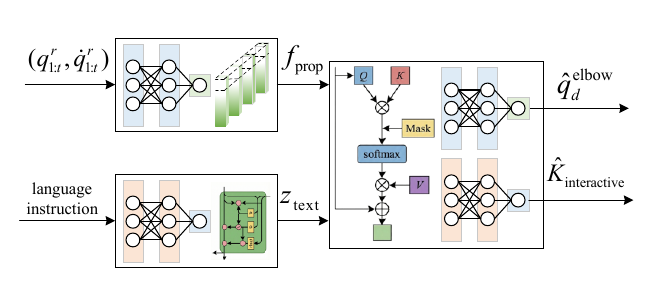}
	\caption{Overall architecture of the bimodal multitask assistance policy.}
	\label{fig_3}
\end{figure}

Following the generation of the comprehensive interaction dataset, a unified multitask exoskeleton assistance policy is trained to predict the subsequent exoskeleton reference motion and interaction stiffness based on semantic language descriptions and historical proprioceptive data $(q^r_{1:t}, \dot{q}^r_{1:t})$. The overall architecture of the proposed neural network is depicted in Fig. \ref{fig_3}. The historical joint positions and velocities are concatenated and linearly projected into a high-dimensional latent space. The resulting features are then encoded via 1D dilated convolutions to extract temporal dependencies.

\begin{figure*}[!t]
	\centering
	\includegraphics[width=\textwidth]{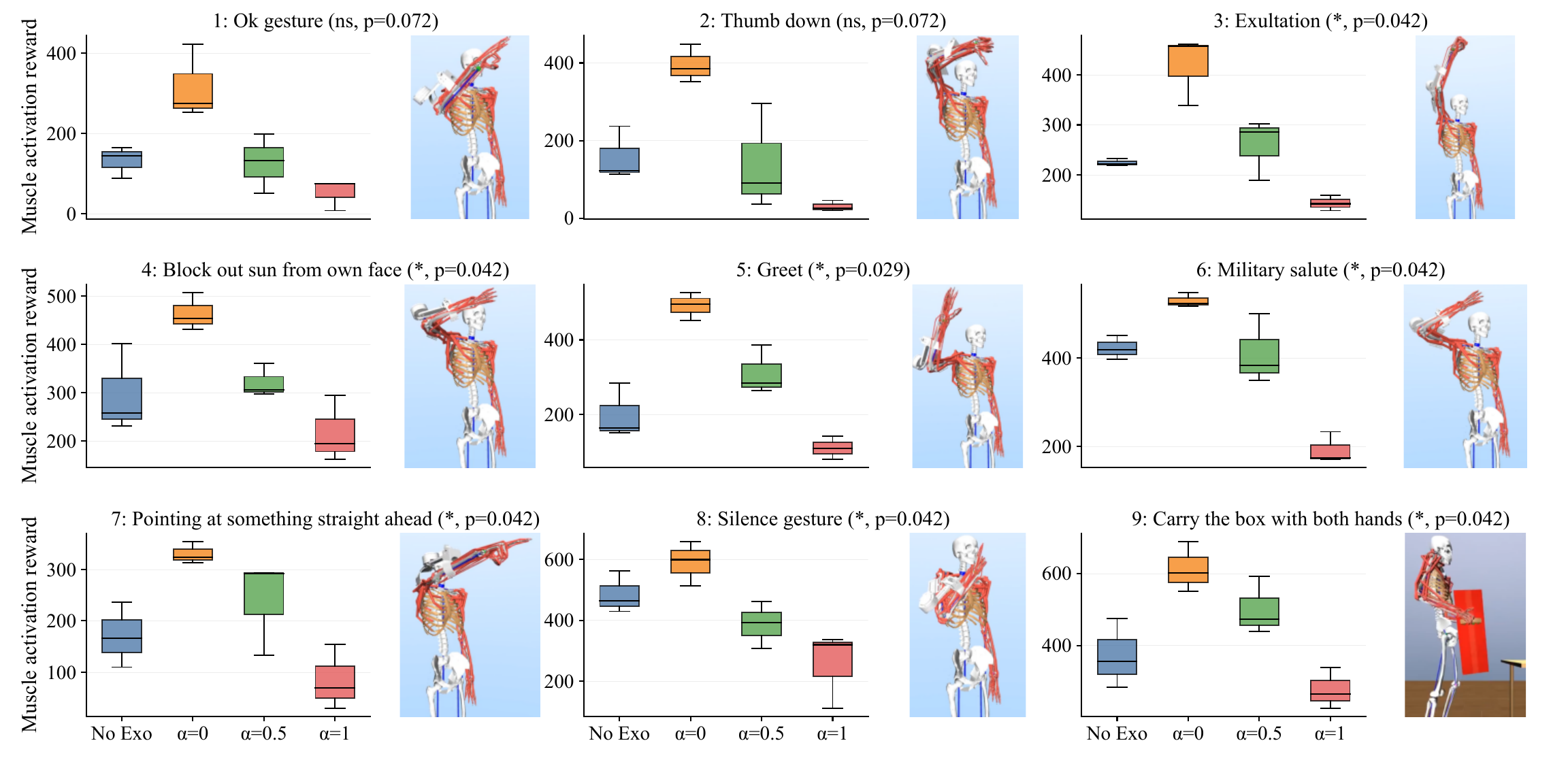}
	\caption{Comparison of muscle activation rewards across nine different simulated tasks under four assistance conditions: without exoskeleton (No Exo), zero assistance ($\alpha=0$), partial assistance ($\alpha=0.5$), and full assistance ($\alpha=1$). Statistical significance from the Friedman test is annotated above each subplot (*: $p < 0.05$, ns: not significant).}
	\label{fig_4}
\end{figure*}

Concurrently, the natural language instruction indicating the task intent is tokenized and processed through an embedding layer and a recurrent neural network. This operation maps the text into dense semantic vectors $z_{\text{text}}$, which share the same high-dimensional latent space as the encoded proprioceptive features $f_{\text{prop}}$. To achieve bimodal fusion, a multi-head cross-attention mechanism is applied to $f_{\text{prop}}$ and $z_{\text{text}}$. In this architecture, $f_{\text{prop}}$ act as the queries, while $z_{\text{text}}$ serve as the keys and values, enabling the policy to dynamically attend to relevant semantic tokens based on the current physical state. 

The feature vector from the final time step of this fused sequence is then fed into two decoupled output heads. To address the different physical characteristics of kinematics and impedance, a linear projection head is employed to directly predict $q^{\text{elbow}}_d$, while a deeper multi-layer perceptron predicts $K_{\text{interactive}}$.
The overall optimization objective is formulated as a weighted sum of the squared $L_2$ norm errors from both prediction heads:
\begin{equation}
	L = \left\| \hat{q}^{\text{elbow}}_d - q^{\text{elbow}}_d \right\|^2 + w_1 \left\| \hat{K}_{\text{interactive}} - K_{\text{interactive}} \right\|^2,
\end{equation}
where $\hat{q}^{\text{elbow}}_d$ and $\hat{K}_{\text{interactive}}$ denote the predicted reference elbow joint position and interaction stiffness, respectively. $w_1$ is a weighting coefficient to balance the numerical scale and gradient contribution of the impedance learning relative to the kinematic tracking.
 
\section{Stability-Guaranteed Variable Impedance Control}
To ensure safe and compliant physical human-robot interaction, a variable impedance controller is designed to track the reference kinematics while dynamically modulating the interactive behavior according to the predicted interaction stiffness. The standard rigid-body dynamics of the exoskeleton in the joint space can be formulated as:
\begin{equation}\label{eq:6}
M(q^r)\ddot{q}^r + C(q^r,\dot{q}^r)\dot{q}^r + g(q^r) = \tau + \tau_{\text{ext}},
\end{equation}
where $\dot{q}^r, \ddot{q}^r$ denote the joint velocity and acceleration vectors, respectively. $M(q^r)$ is the symmetric positive-definite inertia matrix, $C(q^r,\dot{q}^r)$ represents the Coriolis and centrifugal matrix, $g(q^r)$ is the gravity vector, and $\tau$ is the commanded control torque applied by the joint actuators.

The fundamental objective of impedance control is to enforce a desired dynamic relationship between the motion tracking error and the external interaction torque, rather than directly regulating the force or rigidly tracking the trajectory. Let $e = \hat{q}^{\text{elbow}}_d - q^r$ be the tracking error, the target impedance behavior is specified by a second-order linear differential equation:
\begin{equation}\label{eq:7}
	M \ddot{e} + D \dot{e} + \hat{K}_{\text{interactive}} e = \tau_{\text{ext}},
\end{equation}
where $M$ and $D$ are the predefined inertia and damping matrices, respectively.

To transform the nonlinear robot dynamics into this target linear impedance model, an inverse dynamics control law is designed as follows:
\begin{equation}\label{eq:8}
	\begin{split}
		\tau =&~ C(q^r,\dot{q}^r)\dot{q}^r + g(q^r) + M(q^r)M^{-1}(D\dot{e} + \hat{K}_{\text{interactive}}e) \\
		&- \left(I + M(q^r)M^{-1}\right)\tau_{\text{ext}}.
	\end{split}
\end{equation}
By substituting \eqref{eq:8} into \eqref{eq:6}, the inherent nonlinearities are exactly decoupled and canceled, yielding the desired stable impedance behavior \eqref{eq:7}.

\begin{table*}[ht]
	\centering
	\caption{Trajectory Tracking Rewards Across Missions and Assistance Conditions}
	\label{tab_1}
	\footnotesize 
	\renewcommand{\arraystretch}{1.2}
	\setlength{\tabcolsep}{8pt} 
	\resizebox{\textwidth}{!}{
		\begin{tabular}{l ccccccccc}
			\toprule
			\multirow{2}{*}{Condition} & \multicolumn{9}{c}{Mission ID} \\
			\cmidrule(lr){2-10}
			& 1 & 2 & 3 & 4 & 5 & 6 & 7 & 8 & 9 \\
			\midrule
			No Exo 
			& $95 \pm 8$ & $237 \pm 31$ & $135 \pm 6$ & $190 \pm 33$ & $238 \pm 11$ & $220 \pm 38$ & $122 \pm 22$ & $234 \pm 16$ & $406 \pm 77$ \\
			
			$\alpha=0$ 
			& $92 \pm 5$ & $226 \pm 17$ & $132 \pm 2$ & $192 \pm 39$ & $276 \pm 62$ & $172 \pm 14$ & $113 \pm 15$ & $247 \pm 18$ & $371 \pm 42$ \\
			
			$\alpha=0.5$ 
			& $101 \pm 11$ & $261 \pm 51$ & $134 \pm 5$ & $209 \pm 22$ & $236 \pm 30$ & $255 \pm 95$ & $110 \pm 14$ & $243 \pm 39$ & $426 \pm 56$ \\
			
			$\alpha=1$ 
			& $123 \pm 44$ & $229 \pm 8$ & $136 \pm 4$ & $211 \pm 22$ & $238 \pm 22$ & $201 \pm 20$ & $112 \pm 12$ & $238 \pm 27$ & $420 \pm 54$ \\
			
			\midrule
			$p$-value 
			& $0.33$ & $0.46$ & $0.80$ & $0.61$ & $0.90$ & $0.61$ & $0.53$ & $0.80$ & $0.61$ \\
			\bottomrule
			
			\multicolumn{10}{l}{\parbox{16cm}{\vspace{2pt} \footnotesize\textit{Note:} Data are presented as mean $\pm$ standard deviation. Mission-level significance is assessed using the Friedman test with mass as paired blocks; all tests indicate no significant difference ($p > 0.05$).}}
	\end{tabular}}
\end{table*}

However, since $\hat{K}_{\text{interactive}}$ is dynamically predicted by the neural network, rapid variations in the stiffness parameter can inject unbounded energy into the coupled human-robot system, potentially violating passivity and compromising interaction stability. To guarantee the stability of the closed-loop system under the variable impedance control, a stability analysis is conducted. According to \cite{ref26}, consider the following Lyapunov candidate function:
\begin{equation}\label{eq:9}
	V(e, \dot{e}, t) = \frac{1}{2}(\dot{e} + \gamma e)^T M (\dot{e} + \gamma e) + \frac{1}{2} e^T \beta(t) e.
\end{equation}
To ensure that $V(e, \dot{e}, t)$ qualifies as a valid positive definite Lyapunov function, the parameter $\gamma$ is a strictly positive constant selected such that the matrix
\begin{equation}\label{eq:10}
\beta(t) = \hat{K}_{\text{interactive}}(t) + \gamma D - \gamma^2 M
\end{equation}
remains positive semidefinite for all $t > 0$. By differentiating $V$ with respect to time along the closed-loop system trajectories, $\dot{V}(e, \dot{e}, t) \leq 0$ is guaranteed if the following two conditions are simultaneously satisfied:
\begin{equation}\label{eq:cond1}
	\gamma M - D \leq 0,
\end{equation}
\begin{equation}\label{eq:cond2}
	\dot{\hat{K}}_{\text{interactive}}(t) - 2\gamma \hat{K}_{\text{interactive}}(t) \leq 0,
\end{equation}
where the matrix inequalities denote negative semidefiniteness. Furthermore, under these conditions, $\dot{V}(e, \dot{e}, t) = 0$ holds if and only if $e = 0$ and $\dot{e} = 0$. According to LaSalle's invariance principle, the asymptotic stability of the closed-loop system is guaranteed, thereby ensuring safe physical human-exoskeleton interaction. The complete stability proof of the closed-loop system is provided in the Appendix. 

For the 1-DoF elbow exoskeleton, the stability matrices reduce to positive scalars. By setting $\gamma = D/M$ to maximize the allowable stiffness variation rate, the stability condition simplifies to an explicit upper bound for the stiffness derivative:

\begin{equation}\label{eq:k_interactive}
	\dot{\hat{K}}_{\text{interactive}}(t) \leq 2 \frac{D}{M} \hat{K}_{\text{interactive}}(t).
\end{equation}

Physically, \eqref{eq:k_interactive} implies that the exoskeleton can instantaneously decrease its stiffness to become compliant, but its rate of increase is strictly bounded to prevent sudden energy injections. To practically enforce this stability criterion, an asymmetric rate limiter bounded by \eqref{eq:k_interactive} is applied to the network predictions.

\section{Experiment Results}
This section presents a comprehensive evaluation of the proposed framework, divided into simulation analysis and real-world deployment. The first part validates the rationality and kinematic consistency of the synthesized data across nine tasks. Second, real-world deployments are conducted to assess the control stability and multitask adaptability of the simulation-trained variable impedance control approach. Within this real-world phase, the framework's assistive efficacy is further quantified through metabolic cost analysis during a repetitive load-carrying task. All experiment procedures have been reviewed and approved by the ethics committee of Institute of Automation, Chinese Academy of Sciences under the protocol no. IA21-2302-140202.

\subsection{Simulation Results Validation}

\begin{figure*}[ht]
	\centering
	\includegraphics[width=\textwidth]{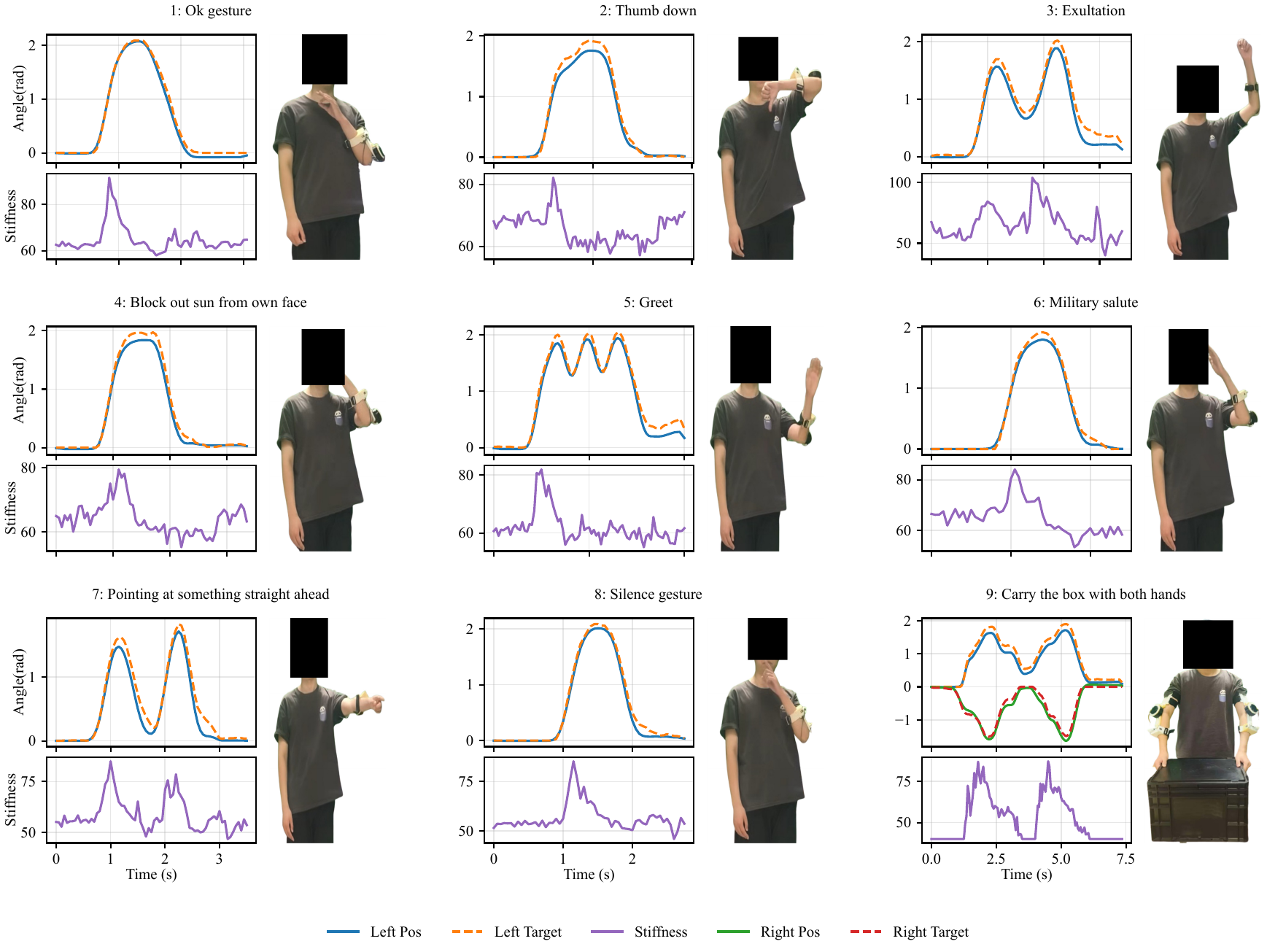}
	\caption{Reference trajectory tracking and dynamically generated interaction stiffness across the nine evaluated tasks in real-world experiments.}
	\label{fig_5}
\end{figure*}

To comprehensively validate the rationality and biomechanical fidelity of the generated data, the variations in elbow-related muscle activation rewards across nine different missions are evaluated. As illustrated in Fig. \ref{fig_4}, the evaluation is conducted under four conditions: without the exoskeleton (No Exo), zero assistance ($\alpha=0$), partial assistance ($\alpha=0.5$), and full assistance ($\alpha=1$), with the results averaged across different musculoskeletal model masses. A lower value of this metric indicates that less muscle activation is required by the musculoskeletal model to drive the joints and accomplish the identical kinematic task. As observed from the Fig. \ref{fig_4}, when the exoskeleton assistance ratio is set to $\alpha=0$, the muscle activation reaches its maximum. This is biomechanically intuitive, as the simulated human model must generate additional effort to overcome the inherent mass and inertia of the unpowered exoskeleton. Conversely, under the full assistance condition ($\alpha=1$), the muscle activation is minimized because the exoskeleton theoretically compensates for the entire human biological torque. However, it is noteworthy that the musculoskeletal model still exhibits residual activation even at $\alpha=1$. This phenomenon occurs because the assistive torque generated by the exoskeleton cannot be losslessly transmitted to the human biological elbow joint, reflecting the realistic compliance and coupling dynamics at the physical human-robot interface. Importantly, the progressive decrease in muscle activation corresponding to the incremental increase in the assistance ratio ($\alpha$) is highly consistent with empirical physiological trends observed in real-world exoskeleton  assistance experiments.

Furthermore, to statistically evaluate the differences among the assistance conditions across various tasks, a Friedman test is conducted for each mission, treating the three musculoskeletal model masses (60 kg, 70 kg, and 80 kg) as paired blocks. The statistical results indicate that there are significant differences ($p < 0.05$) in muscle activation for seven out of the nine evaluated missions. The only exceptions are the ``Ok gesture'' and ``Thumb down'' tasks, which do not reach statistical significance. To ensure that the observed reduction in muscle effort does not compromise task execution, an identical statistical analysis is conducted on the trajectory tracking error rewards. The statistical results of the trajectory tracking rewards are summarized in Table \ref{tab_1}, which reveals no significant differences across all evaluated missions ($p > 0.05$). This demonstrates that the varying levels of exoskeleton assistance successfully reduce the required muscle effort on the user without sacrificing the kinematic tracking accuracy of the tasks.

\subsection{Real-World Exoskeleton Experiments}

Experiments are conducted on an elbow exoskeleton \cite{ref27} equipped with Series Elastic Actuators (SEAs) for force sensing at the human-robot interface. To evaluate the proposed approach, comparative experiments are conducted across the nine missions from the dataset under the following three conditions:
\begin{enumerate}
	\item \textbf{Natural Movement:} Task execution without the exoskeleton.
	\item \textbf{Simulation-Trained Variable Impedance Approach:} The proposed framework designed to provide stable assistance.
	\item \textbf{Probabilistic Movement Primitives (ProMP) \cite{ref28}:} A standard data-driven baseline employed to generate reference joint trajectories and interaction stiffness.
\end{enumerate}
As demonstrated in the supplementary video, the real-world evaluation comprises two primary phases: kinematic tracking across nine different missions and a physically demanding repetitive load-carrying metabolic assessment.

The nine missions are executed to evaluate the framework's control stability and multitask adaptability. Fig. \ref{fig_5} presents the reference and actual joint trajectories alongside the corresponding interaction stiffness profiles across the nine missions. Based on the kinematic profiles, the evaluated tasks can be categorized into three distinct patterns: missions 1, 2, 4, 6, and 8 exhibit single-peak bell-shaped trajectories; missions 3, 7, and 9 display double-peak curves; and mission 5 presents a triple-peak pattern. Moreover, a consistent correlation is observed regarding the dynamic impedance modulation, whereby the interaction stiffness significantly increases only during the phases of rapid joint angle increase. In contrast, the stiffness generally stabilizes or decreases during the steady-state or descending phases of the joint angle. Specifically for the bimanual carrying task (Mission 9), the interaction stiffness is unified to ensure symmetric assistive torque and maintain payload stability during execution:
\begin{equation}
	\hat{K}_{\text{interactive}}(t) = \frac{\hat{K}_{\text{interactive}}^{\text{right}}(t) + \hat{K}_{\text{interactive}}^{\text{left}}(t)}{2}.
\end{equation}

\begin{figure}[ht]
	\centering
	\includegraphics[width=\columnwidth]{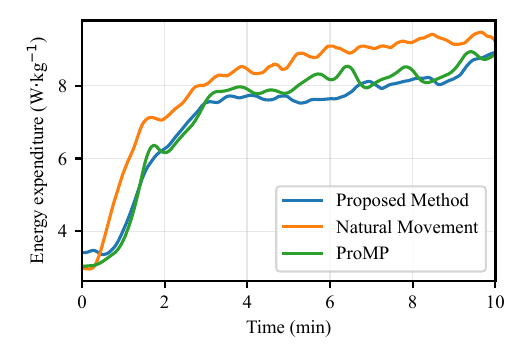}
	\caption{Trajectories of mass-normalized metabolic energy expenditure ($W \cdot kg^{-1}$) during the 10-minute task. The curves compare the assistive performance of the proposed method against natural movement and the baseline ProMP condition.}
	\label{fig_6}
\end{figure}

The root-mean-square error (RMSE) serves as a comprehensive system-level metric for quantifying performance, capturing the coupled impact of reference trajectory prediction errors from the learned policy and tracking errors from the low-level controller. Across the nine missions, the overall average RMSE is 0.1 $\pm$ 0.03 rad. The maximum tracking error of 0.14 rad occurs during the third mission, while the minimum error of 0.06 rad is observed in the sixth mission. 

Furthermore, to quantify the assistive efficacy of the exoskeleton, the respiratory metabolic cost is measured during a 10-minute repetitive 10 kg lifting and carrying task, which represents the most physically demanding activity within the dataset. Metabolic data is recorded via indirect calorimetry using a portable analyzer equipped with a respiratory mask (PACER VO2 Analyzer, Neumafit Co. Ltd., South Korea). The metabolic energy expenditure ($W \cdot kg^{-1}$) is computed from the raw volumetric rates of oxygen consumption  $\dot{V}_{O_2}$ and carbon dioxide production $\dot{V}_{CO_2}$ using the Brockway equation:
\begin{equation}
EE = \frac{16.58 \dot{V}_{O_2} + 4.51 \dot{V}_{CO_2}}{m},
\end{equation}
where $m$ denotes the subject's body mass. During the 10-
minute trial, the subject completes 67 box transfers under
natural movement, 62 transfers under the ProMP-assisted
condition, and 64 transfers using the proposed assistive
method. This suggests that the effective workload remains
broadly comparable across conditions.

The variations in energy expenditure across the three experimental conditions are illustrated in Fig. \ref{fig_6}. To completely eliminate any potential fatigue carry-over, the three experimental conditions are evaluated on separate days, with a strict 24-hour rest period enforced between sessions. The trajectories show a transient increase in metabolic energy expenditure during the initial 3 minutes of the task due to physiological adaptation to the physical demand. The energy expenditure subsequently stabilizes and enters a steady-state phase from 3 to 10 minutes. The total steady-state energy consumption ($EC$) is calculated based on these measurements using the equation:
\begin{equation}
	EC = m \int_{t_1}^{t_2} EE(t) dt,
\end{equation}
where $[t_1, t_2]$ denotes the steady-state time window in seconds. The results indicate a total steady-state energy consumption of 250.2 kJ for natural movement. Relative to this baseline, the proposed method achieves a 10.9\% reduction (consuming  223.0 kJ), whereas the ProMP framework yields a smaller reduction of 8.0\% (230.1 kJ).
These findings suggest the potential of the proposed framework to mitigate human fatigue during repetitive load-carrying activities.

\begin{figure}[ht]
	\centering
	\includegraphics[width=\columnwidth]{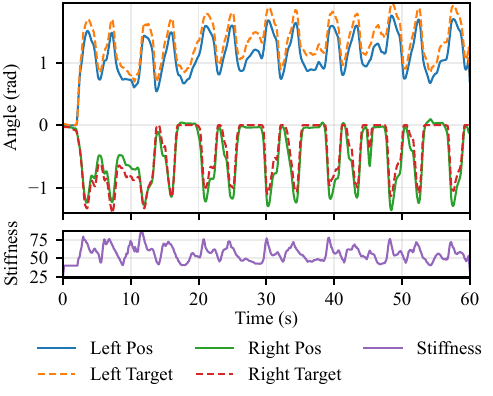}
	\caption{Kinematic tracking performance and variable stiffness modulation during the initial 1 minute of the repetitive 10 kg lifting task.}
	\label{fig_7}
\end{figure}

Beyond the metabolic benefits, the joint-level kinematic tracking performance during the repetitive task is also evaluated. The RMSE of the joint angle is calculated as 0.2 rad. This tracking error is larger than the RMSE observed in the isolated single-lifting task. The increased deviation is attributed to severe dynamic disturbances induced by continuous 10 kg payload manipulation and high-frequency task transitions. The tracking trajectories and the corresponding stiffness modulation during the initial 1 minute of the trial are detailed in Fig. \ref{fig_7}. As depicted, the stiffness increases when the joint angle rises rapidly, which is consistent with the isolated mission.

\section{Conclusions}
This paper proposes a simulation-trained variable impedance control approach for wearable exoskeletons that enables safe and multitask assistance. The proposed dual modality policy uses data generated in simulation to predict task-specific trajectories and stiffness while ensuring safe human-robot interaction through rigorous Lyapunov-based stability bounds. Real-world experiments validate the efficacy of the proposed framework, demonstrating effective multitask assistance and reductions in user metabolic cost compared to standard baselines during repetitive load-carrying tasks. Future work will focus on extending the proposed framework to multi-joint exoskeletons and evaluating its efficacy across a broader range of participants.

\section*{Appendix \\ Proof of Closed-Loop System Stability}
This Appendix describes the proof of the asymptotic stability for the closed-loop human-exoskeleton system. The foundational theoretical framework and the stability criteria follow the approach established by \cite{ref26}. Taking the time derivative of \eqref{eq:9} yields: 
\begin{equation}\label{eq:v_dot1}
	\begin{split}
		\dot{V}(e, \dot{e}, t) =&~ (\dot{e} + \gamma e)^T M (\ddot{e} + \gamma \dot{e}) 
		+ e^T \beta(t) \dot{e} \\
		&+ \frac{1}{2} e^T \dot{\beta}(t) e.
	\end{split}
\end{equation}
In the absence of $\tau_{\text{ext}}$, substituting \eqref{eq:7} into \eqref{eq:v_dot1} results in:
\begin{equation}\label{eq:v_dot2}
	\begin{split}
		\dot{V}(e, \dot{e}, t) =&~ \dot{e}^T ( \gamma M - D ) \dot{e} \\
		&+ \dot{e}^T \{ \beta(t) + \gamma^2 M - \hat{K}_{\text{interactive}}(t) - \gamma D \} e \\
		&+ e^T \left\{ \frac{1}{2}\dot{\beta}(t) - \gamma \hat{K}_{\text{interactive}}(t) \right\} e.
	\end{split}
\end{equation}
According to \eqref{eq:10}, the cross-coupling term evaluates to zero. With the property $\dot{\beta}(t) = \dot{\hat{K}}_{\text{interactive}}(t)$, \eqref{eq:v_dot2} reduces to:

\begin{equation}\label{eq:v_dot3}
	\begin{split}
		\dot{V}(e, \dot{e}, t) =&~ \frac{1}{2} e^T \left\{ \dot{\hat{K}}_{\text{interactive}}(t) - 2\gamma \hat{K}_{\text{interactive}}(t) \right\} e \\
		&+ \dot{e}^T ( \gamma M - D ) \dot{e}.
	\end{split}
\end{equation}
Therefore, provided that conditions \eqref{eq:cond1} and \eqref{eq:cond2} are satisfied, it is mathematically guaranteed that $\dot{V}(e, \dot{e}, t) \leq 0$. This concludes the proof.

\addtolength{\textheight}{-1cm}


\begin{thebibliography}{99}
\bibitem{1}
R. Bogue, ``Robotic exoskeletons: A review of recent progress,'' \textit{Industrial Robot: An International Journal}, vol. 42, no. 1, pp. 5--10, 2015.

\bibitem{2}
A. Carnevale, G. Nicodemi, M. G. Pisani, A. Lalli, F. S. di Luzio, 
P. D'Hooghe, L. Zollo, E. Schena, and U. G. Longo,
``Portable exoskeletons for upper limb rehabilitation: A systematic review,''
\emph{Journal of Experimental Orthopaedics},
vol. 12, no. 3, art. no. e70416, 2025.

\bibitem{3}
G. Bao, L. Pan, H. Fang, X. Wu, H. Yu, S. Cai, B. Yu, and Y. Wan,
``Academic review and perspectives on robotic exoskeletons,''
\emph{IEEE Transactions on Neural Systems and Rehabilitation Engineering},
vol. 27, no. 11, pp. 2294--2304, 2019.	

\bibitem{1.1}
P. Sedighi, X. Li, and M. Tavakoli,
``EMG-based intention detection using deep learning for shared control in upper-limb assistive exoskeletons,''
\emph{IEEE Robotics and Automation Letters},
vol. 9, no. 1, pp. 41--48, 2024.

\bibitem{1.3}
H. Li, L. Cheng, S. Qin, and L. Han,
``Voluntary control of the hand assistive exoskeleton based on the sEMG-driven musculoskeletal model,''
\emph{IEEE Robotics and Automation Letters}, vol. 10, no. 7, pp. 7651--7658, 2025.


\bibitem{1.10}
G. Rinaldi, V. Suglia, L. Tiseni, C. Camardella, M. Xiloyannis, 
L. Masia, D. Buongiorno, V. Bevilacqua, A. Frisoli, and D. Chiaradia,
``Towards a healthier workplace: How Flexos, an active and bilateral shoulder exoskeleton, provides support in weight-lifting and carrying tasks,''
\emph{IEEE Transactions on Robotics}, in press, DOI: 10.1109/TRO.2026.3666155.

\bibitem{1.7}
B. Luciani, L. Roveda, F. Braghin, A. Pedrocchi, and M. Gandolla, ``Trajectory learning by therapists' demonstrations for an upper limb rehabilitation exoskeleton,''
\emph{IEEE Robotics and Automation Letters}, vol. 8, no. 8, pp. 4561--4568, 2023.

\bibitem{2.6}
J. Lee, K. Kwon, I. Soltis, J. Matthews, Y. J. Lee, H. Kim, L. Romero, N. Zavanelli, 
Y. Kwon, S. Kwon, J. Lee, Y. Na, S. H. Lee, K. J. Yu, M. Shinohara, F. L. Hammond, and W.-H. Yeo,
``Intelligent upper-limb exoskeleton integrated with soft bioelectronics and deep learning for intention-driven augmentation,''
\emph{npj Flexible Electronics},
vol. 8, no. 1, art. no. 11, 2024.

\bibitem{1.8}
L. D. Arco, K. Gusakowski, C. A. Cifuentes, M. Munera, M. Segatto, and C. A. R. Díaz, 
``Mitigating muscle fatigue in upper-limb prosthesis users through exoskeletal weight compensation,'' 
\emph{IEEE Transactions on Neural Systems and Rehabilitation Engineering}, vol. 34, pp. 334--344, 2026.

\bibitem{3.2}
D. D. Molinaro, K. L. Scherpereel, E. B. Schonhaut, G. Evangelopoulos, M. K. Shepherd, and A. J. Young,
``Task-agnostic exoskeleton control via biological joint moment estimation,''
\emph{Nature}, 
vol. 635, pp. 337--344, 2024.

\bibitem{2.2}
H. Wang, Y. Yao, H. Lei, Y. Shi, and S. Pei, ``High-accuracy early recognition of upper-limb motions for exoskeleton-assisted mirror rehabilitation,''
\emph{IEEE Robotics and Automation Letters},
vol. 10, no. 3, pp. 2718--2725, 2025.

\bibitem{2.9}
Y. Zimmermann, M. Sommerhalder, P. Wolf, R. Riener, and M. Hutter, 
``ANYexo 2.0: A fully actuated upper-limb exoskeleton for manipulation and joint-oriented training in all stages of rehabilitation,'' 
\emph{IEEE Transactions on Robotics}, 
vol. 39, no. 3, pp. 2131--2150, 2023.

\bibitem{2.10}
Y. Chen, S. Miao, G. Chen, J. Ye, C. Fu, B. Liang, S. Song, and X. Li,
``Learning to assist different wearers in multitasks: efficient and individualized human-in-the-loop adaptation framework for lower-limb exoskeleton,''
\emph{IEEE Transactions on Robotics}, 
vol. 40, pp. 4699--4718, 2024.

\bibitem{myo}
H. Wang, V. Caggiano, G. Durandau, M. Sartori, and V. Kumar,
``MyoSim: Fast and physiologically realistic MuJoCo models for musculoskeletal and exoskeletal studies,''
in \emph{Proceedings of the 38th International Conference on Robotics and Automation}, Philadelphia, USA, 2022, pp. 8104--8111.

\bibitem{3.3.3}
N.~Wilhelm, V.~Schaack, A.~Leisching, C.~Micheler, S.~Haddadin, and R.~Burgkart,
``An adaptive robotic exoskeleton for comprehensive force-controlled hand rehabilitation,''
in \textit{Proceedings of the 36th IEEE/RSJ International Conference on Intelligent Robots and Systems},
Abu Dhabi, United Arab Emirates, 2024, pp.~170--177.

\bibitem{robustwalking}
S. Luo, G. Androwis, S. Adamovich, E. Nunez, H. Su, and X. Zhou,
``Robust walking control of a lower limb rehabilitation exoskeleton coupled with a musculoskeletal model via deep reinforcement learning,''
\emph{Journal of NeuroEngineering and Rehabilitation},
vol. 20, no. 1, art. no. 34, 2023.

\bibitem{3.1}
S.~Luo, M.~Jiang, S.~Zhang, J.~Zhu, S.~Yu, I.~Dominguez~Silva, T.~Wang, E.~Rouse, B.~Zhou, H.~Yuk, X.~Zhou, and H.~Su,
``Experiment-free exoskeleton assistance via learning in simulation,''
\emph{Nature},
vol. 630, pp. 353--359, 2024.

\bibitem{3.3} K. L. Scherpereel, M. C. Gombolay, M. K. Shepherd, C. A. Carrasquillo, O. T. Inan, and A. J. Young, ``Deep domain adaptation eliminates costly data required for task-agnostic wearable robotic control,'' \emph{Science Robotics}, vol. 10, no. 108, art. no. eads8652, 2025.


\bibitem{ref25}
G. Averta, F. Barontini, V. Catrambone, S. Haddadin, G. Handjaras, J. P. Held, T. Hu, E. Jakubowitz, C. M. Kanzler, J. Kühn, O. Lambercy, A. Leo, A. Obermeier, E. Ricciardi, A. Schwarz, G. Valenza, A. Bicchi, and M. Bianchi, ``U-Limb: A multi-modal, multi-center database on arm motion control in healthy and post-stroke conditions,'' \textit{GigaScience}, vol. 10, no. 6, art. no. giab043, 2021.

\bibitem{ref25-a}
Carnegie Mellon University, ``CMU Graphics Lab Motion Capture Database,'' [Online]. Available: http://mocap.cs.cmu.edu/

\bibitem{ref26}
K. Kronander and A. Billard, ``Stability considerations for variable impedance control,'' \textit{IEEE Transactions on Robotics}, vol. 32, no. 5, pp. 1298--1305, 2016.

\bibitem{ref27}
X. Xia, L. Han, H. Li, Y. Zhang, Z. Liu, and L. Cheng, ``A compliant elbow exoskeleton with an SEA at interaction port,'' \textit{in Proceedings of the 30th International Conference on Neural Information Processing}, Changsha, China, 2024, pp. 146--157.

\bibitem{ref28}
Y. Chen, G. Chen, J. Ye, X. Qiu, and X. Li, ``Safe and individualized motion planning for upper-limb exoskeleton robots using human demonstration and interactive learning,'' \textit{in Proceedings of the 40th International Conference on Robotics and Automation}, Yokohama, Japan, 2024, pp. 15307--15313.

\end{thebibliography}
\end{document}